\documentclass[10pt,twocolumn,letterpaper]{article}

\usepackage[pagenumbers]{cvpr} % To force page numbers, e.g. for an arXiv version

\usepackage{bm}

\definecolor{cvprblue}{rgb}{0.21,0.49,0.74}
\usepackage[pagebackref,breaklinks,colorlinks,allcolors=cvprblue]{hyperref}
\usepackage{multirow}

\title{SOGS: Second-Order Anchor for Advanced 3D Gaussian Splatting}

\author{Jiahui Zhang\textsuperscript{\rm 1}
\quad Fangneng Zhan\textsuperscript{\rm 2, 3} 
\quad Ling Shao\textsuperscript{\rm 4} 
\quad Shijian Lu\textsuperscript{\rm 1} \\
{ $^1$Nanyang Technological University\quad$^2$Harvard University \quad$^3$MIT}\\[0.1mm]
{ $^4$UCAS-Terminus AI Lab, University of Chinese Academy of Sciences} \\
{\tt\small \ jiahui003@e.ntu.edu.sg \ fnzhan@seas.harvard.edu} \\
{\tt\small\ ling.shao@ieee.org \ shijian.lu@ntu.edu.sg }
}

\begin{document}

\twocolumn[{
    \renewcommand\twocolumn[1][]{#1}
    \maketitle
    \begin{center}
        \captionsetup{type=figure}
        \includegraphics[width=1.\linewidth]{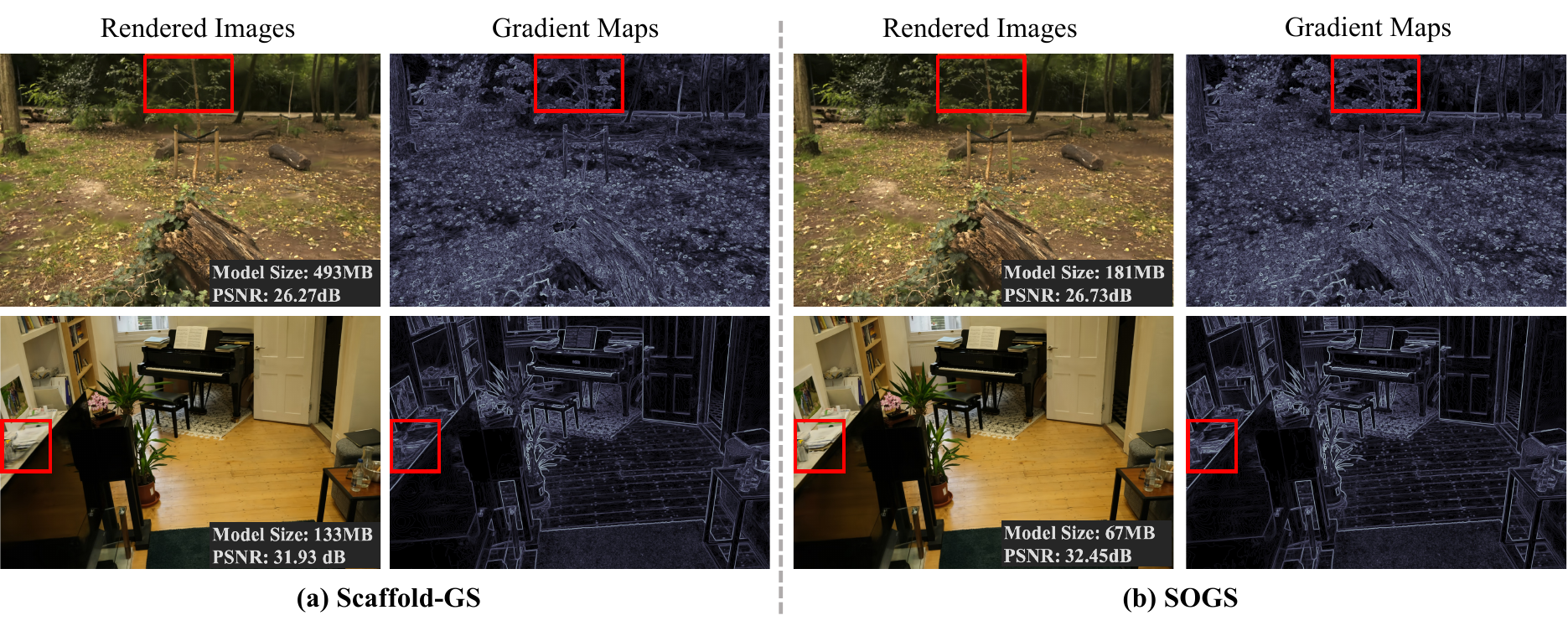}
        \captionof{figure}{
        The proposed SOGS can render high-quality textures and geometries and reduce model size simultaneously. The illustrations in (a) and (b) show \textit{Rendered Images} and \textit{Gradient Maps} which are produced by Scaffold-GS~\cite{lu2024scaffold} and SOGS, respectively, for the samples `Stump' and `Room' from Mip-NeRF360~\cite{barron2022mip}. The \textit{Gradient Maps} are extracted with the Sobel operator which highlight the texture and geometry of imaged scenes.
        }
        \label{teaser}
    \end{center}

}]
\def\thefootnote{*}\footnotetext{Shijian Lu is the corresponding author.}

% \maketitle
\begin{abstract}
Anchor-based 3D Gaussian splatting (3D-GS) exploits anchor features in 3D Gaussian prediction, which has achieved impressive 3D rendering quality with reduced Gaussian redundancy. On the other hand, it often encounters the dilemma among anchor features, model size, and rendering quality – large anchor features lead to large 3D models and high-quality rendering whereas reducing anchor features degrades Gaussian attribute prediction which leads to clear artifacts in the rendered textures and geometries. We design SOGS, an anchor-based 3D-GS technique that introduces second-order anchors to achieve superior rendering quality and reduced anchor features and model size simultaneously. Specifically, SOGS incorporates covariance-based second-order statistics and correlation across feature dimensions to augment features within each anchor, compensating for the reduced feature size and improving rendering quality effectively. In addition, it introduces a selective gradient loss to enhance the optimization of scene textures and scene geometries, leading to high-quality rendering with small anchor features. Extensive experiments over multiple widely adopted benchmarks show that SOGS achieves superior rendering quality in novel view synthesis with clearly reduced model size.
\end{abstract}
    
\section{Introduction}

3D scene representation and rendering have been a pivotal task in 3D computer vision, playing a crucial role in various applications such as virtual reality and scene simulation. Neural radiance fields (NeRFs)~\cite{mildenhall2020nerf} and its variants~\cite{zhang2020nerf++, barron2021mip, barron2022mip, barron2023zip} have been developed to model implicit scene representation and enable volume rendering for high-quality novel view synthesis. However, NeRF is limited by long training and rendering times, largely due to the time-consuming point sampling in volume rendering. Recently, 3D Gaussian Splatting (3D-GS)~\cite{kerbl20233d} has been proposed as an alternative which learns explicit 3D scene representations with learnable 3D Gaussians. By leveraging efficient splatting and rasterization, 3D-GS projects 3D Gaussians onto a 2D plane, enabling real-time rendering. However, 3D-GS stacks substantial 3D Gaussians to fit each training view individually, often overlooking the underlying scene structure~\cite{lu2024scaffold}.

The recent Scaffold-GS~\cite{lu2024scaffold} exploits scene geometry to guide the distribution of 3D Gaussians, reducing Gaussian redundancy and delivering impressive 3D rendering performance. Specifically, it introduces anchor points to store features and exploits MLPs to predict Gaussian attributes from these anchor features. However, Scaffold-GS struggles to strike a balance between rendering quality and model size. Specifically, adopting large anchor features improves rendering quality~\cite{chen2024hac} but it significantly enlarges model size due to the large number of anchor points used in scene representations. Using smaller anchor features can shrink the model, but it clearly degrades the prediction of Gaussian attributes which further leads to suboptimal rendering with various artifacts in rendered textures and geometries. HAC~\cite{chen2024hac} introduces grid-based context modeling and entropy encoding to compress the Scaffold-GS. However, HAC does not reduce the model size but the storage size of the trained Scaffold-GS. 

We design SOGS, an innovative anchor-based 3D-GS technique that introduces second-order anchors and achieves superior rendering quality with reduced anchor feature dimensions and model size. Specifically, SOGS incorporates covariance-based second-order statistics to model correlation across anchor feature dimensions. 
The rationale is that textures and geometries are defined by not only individual features but also the correlation across features. Hence, the correlation across anchor features can effectively guide to capture the pattern of intricate local textures and structures in each anchor. This nice feature enables anchor feature augmentation, empowering SOGS to achieve superior rendering quality with reduced feature dimensions and feature size.

Additionally, we design a selective gradient loss to enhance the rendering quality with small anchor features. 
This loss computes gradient maps from both the rendered image and the corresponding ground truth that highlight their differences in scene textures and structures. With the difference of the gradient maps, the 3D Gaussian prediction can adaptively focus on image regions where the texture and geometry are hard to render. This enables dynamic region selection along the training process, where the model can adjust the focused image regions dynamically according to the evolving errors between the gradient maps of the rendered image and the ground truth. Experiments show that SOGS simultaneously achieves superior rendering quality in novel view synthesis as well as clearly reduced model sizes as shown in Fig.~\ref{teaser}.

The major contributions of this work can be summarized in three aspects. 
\textit{First}, we propose SOGS, an innovative anchor-based 3D-GS that introduces second-order anchors to balance the rendering quality and model size. 
With covariance-based second-order statistics that model correlation across anchor feature dimensions to capture intricate textural and structural patterns, SOGS achieves anchor feature augmentation which produces superior rendering quality with significantly reduced anchor feature dimensions.
\textit{Second}, 
we design a selective gradient loss that guides 3D Gaussian prediction to focus on difficult-to-render textures and structures adaptively, further enhancing rendering quality with small anchor features
\textit{Third}, experiments over multiple benchmarks show that SOGS achieves superior novel view synthesis with efficient anchor size and model size, and outperforms the state-of-the-art in rendering quality.
\section{Related Work}

\subsection{Radiance Field and Neural Rendering}

Radiance field has been widely explored for novel view synthesis and achieved impressive rendering quality. Neural radiance field (NeRF)~\cite{mildenhall2020nerf} utilizes MLPs to learn implicit 3D scene representations from multi-view posed 2D images. Coupled with differentiable volume rendering, NeRF achieves high-quality rendering with superb multi-view consistency. Several NeRF variants have been developed to address various new challenges, including large-scale scenes~\cite{turki2022mega, tancik2022block, xiangli2022bungeenerf}, dynamic scenes \cite{du2021neural, gao2021dynamic, shao2023tensor4d}, few-shot setting \cite{chen2021mvsnerf, yang2023freenerf, wang2023sparsenerf}, pose-free setting \cite{lin2021barf, bian2023nope, zhang2023pose, chen2023local, zhang2022vmrf}, and antialiasing \cite{barron2022mip, barron2023zip}. However, NeRF and its variants still suffer from extremely long training and rendering times due to the time-consuming sampling required for volume rendering. Several studies~\cite{muller2022instant, garbin2021fastnerf, reiser2021kilonerf, chen2022tensorf} have been proposed to address this issue. For instance, Müller et al. \cite{muller2022instant} introduce multi-resolution hash encoding to reduce the size of neural networks, significantly shortening the training time and enabling real-time rendering in tens of milliseconds. However, these methods often suffer from degraded reconstruction accuracy and rendering quality of scenes. 

Kerbl et al. recently propose 3D Gaussian splatting (3D-GS)~\cite{kerbl20233d}, which achieves high-quality and real-time rendering by explicitly representing scenes by using parameterized 3D Gaussians, along with efficient splatting and rasterization. Building on its excellent performance, numerous 3D-GS variants have been developed to address diverse challenges and tasks, such as dynamic scenes~\cite{li2024spacetime, lin2024gaussian, lu20243d, yang2024deformable, wu20244d},  autonomous driving~\cite{zhou2024drivinggaussian, yan2024street}, rendering quality optimization~\cite{cheng2024gaussianpro, yu2024mip, yan2024multi, zhang2024fregs}, sparse-view setting~\cite{li2024dngaussian, chen2024mvsplat, charatan2024pixelsplat} and colmap-free setting~\cite{fan2024instantsplat}, and 3D style transfer~\cite{liu2024stylegaussian}.

\begin{figure*}[ht]
\begin{center}
\includegraphics[width=1\linewidth]{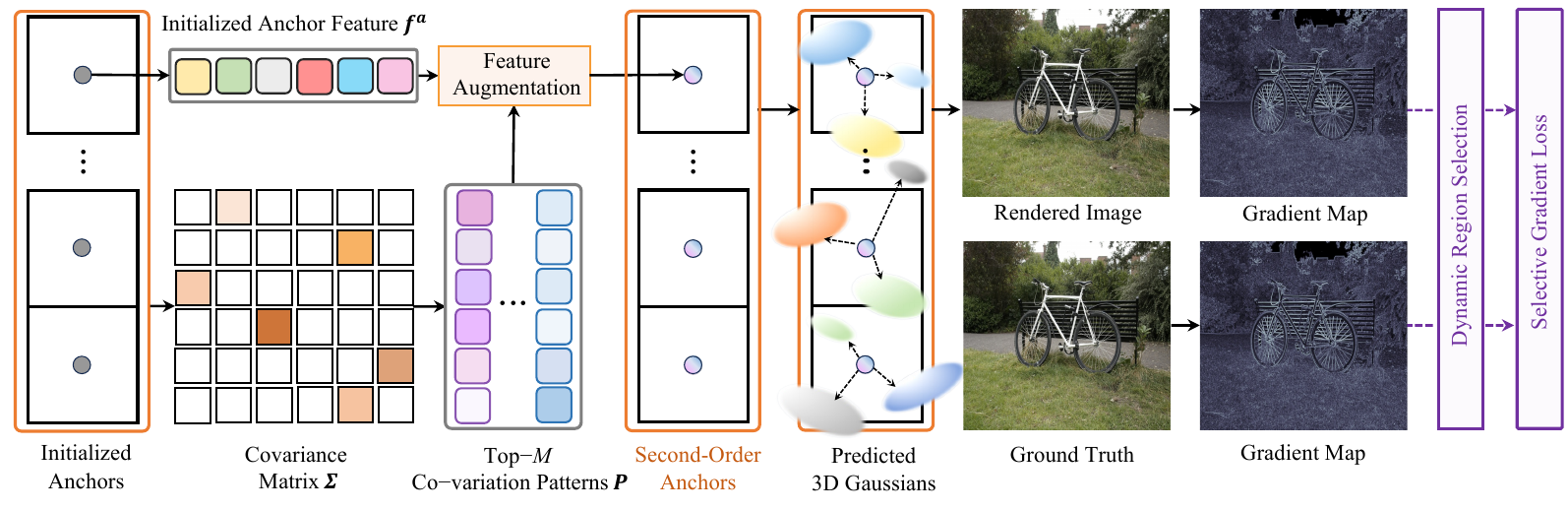}
\end{center}
\caption{
\textbf{Overview of the proposed SOGS.} Initialized from point clouds, each anchor stores an anchor feature vector $\boldsymbol{f^a} \in \mathbb{R}^{D}$. Putting $\boldsymbol{f^a}$ along $D$ dimensions as $D$ variables and all anchors as observed samples, the second-order anchor statistics capturing co-varying relations across $D$ can be computed from the covariance matrix $\boldsymbol{\Sigma}$ across $D$ dimensions. We select the top $M$ eigenvectors $\boldsymbol{P}$ as the most significant co-variation patterns and combine $\boldsymbol{P}$ with $f^a$ to capture anchor-specific textures and geometries for anchor feature augmentation, second-order anchor construction and Gaussian prediction. With the gradient maps of the rendered image and the ground truth, the selective gradient loss guides to learn to render finer textures and geometries with dynamic region selection.}
\label{overview}
\end{figure*}

\subsection{Anchor-based 3D Gaussian Splatting}

One major constraint in 3D-GS is that it stacks a large number of 3D Gaussians to match each training view separately, often overlooking the underlying scene geometry. To address this issue, Lu et al.~\cite{lu2024scaffold} present Scaffold-GS, the first anchor-based 3D-GS technique that incorporates anchor points to predict Gaussian attributes through multi-layer perceptrons (MLPs). Scaffold-GS leverages scene structure to guide the distribution of 3D Gaussians, reducing Gaussian redundancy and delivering impressive 3D rendering performance. However, Scaffold-GS faces challenges in balancing enhanced rendering quality with anchor and model size reduction. Although several follow-ups~\cite{chen2024hac, wang2024contextgs} introduce context-based entropy encoding to compress the Scaffold-GS, they reduce its storage size instead of anchor and model sizes as used during training and rendering. The proposed SOGS introduces second-order anchors, enabling superior rendering quality with reduced model size.
\section{Proposed Method}

We propose SOGS, a novel anchor-based 3D Gaussian splatting that introduces second-order anchors and selective gradient loss to achieve superior rendering quality with reduced anchor and model sizes. Fig.~\ref{overview} shows the overview of SOGS. The 3D Gaussian splatting~\cite{kerbl20233d} (3D-GS) and Scaffold-GS~\cite{lu2024scaffold} (the base model for our SOGS) are briefly introduced in Sec.~\ref{scaffold-gs}. In Sec.~\ref{second-order}, we first explain why the proposed second-order anchor enables the model to enhance rendering quality in reduced anchor and model sizes, and then describe the details of the second-order anchor. Besides, to further ensure high-quality image rendering under compact anchor size, we design a selective gradient loss, detailed in Sec.~\ref{selective_loss}.
\subsection{Preliminaries}
\label{scaffold-gs}

\paragraph{3D Gaussian Splatting (3D-GS).} 3D-GS~\cite{kerbl20233d} explicitly represents scenes using anisotropic 3D Gaussians and enables real-time rendering through efficient differentiable splatting. 3D Gaussians are initialized from point clouds generated by structure-from-motion \cite{hartley2003multiple, schonberger2016structure}. Each Gaussian is defined by a covariance matrix, a position (mean), an opacity value $\alpha$, and spherical harmonics coefficients representing color $c$, where the covariance matrix is decomposed into scaling matrix and rotation matrix to facilitate differentiable optimization.

For rendering, 3D Gaussians are projected onto a 2D plane through splatting, followed by $\alpha$-blending. Specifically, the color $C$ of a pixel is computed by blending $N$ ordered 2D Gaussians overlapping that pixel, formulated as:
\begin{equation}
C = \sum_{n \in N} c_n \alpha_n \prod_{z=1}^{n-1}(1-\alpha_z),
\label{render_formula}
\end{equation}
where $c_n$ and $\alpha_n$ represent the color and opacity of the $n$-th 2D Gaussian.

\paragraph{Scaffold-GS. } Scaffold-GS~\cite{lu2024scaffold} introduces anchor points to predict 3D Gaussians via MLPs. It aims to leverage scene geometry to guide the distribution and attributes of 3D Gaussians and reduce the Gaussian redundancy. Specifically, Scaffold-GS voxelizes the entire scene based on the point cloud initialized by COLMAP~\cite{schonberger2016structure} and assigns the center of each voxel as an anchor point. A local scene feature $\boldsymbol{f^a} \in \mathbb{R}^{D}$, a scaling factor $\boldsymbol{l^a}$ and $K$ offsets $\boldsymbol{o^a} \in \mathbb{R}^{3K}$ are stored in each anchor as the anchor attributes. Given an anchor position $\boldsymbol{x}_a$, its relative distance $\bm{\delta}_{ac}$ and viewing direction $\boldsymbol{d}_{ac}$ to a camera position $\boldsymbol{x}_c$ can be represented as: $\bm{\delta}_{ac} = ||\boldsymbol{x}_a-\boldsymbol{x}_c||_2$, $\boldsymbol{d}_{ac}=\frac{\boldsymbol{x}_a-\boldsymbol{x}_c}{||\boldsymbol{x}_a-\boldsymbol{x}_c||_2}$. 
The prediction of $K$ Gaussians from one anchor point using MLPs $F$ can then be formulated as follows:
\begin{equation}
\{\alpha_k, c_k, q_k, s_k\}_{k=1}^{K} = F(\boldsymbol{f^a}, \bm{\delta}_{ac}, \boldsymbol{d}_{ac}), 
\end{equation}
where $\alpha_k$, $c_k$, $s_k$ and $q_k$ denote the opacity, color, scaling and the rotation-related quaternion of the $k$-th Gaussian, respectively. The positions of the $K$ Gaussians are calculated by this formula:
\begin{equation}
\{p_k\}_{k=1}^{K} = \boldsymbol{x}_a + \{o^a_{k}\}_{k=1}^{K} * \boldsymbol{l^a}.
\end{equation}
Scaffold-GS follows the same rendering process as Eq.~\ref{render_formula} described in 3D-GS after predicting the Gaussian attributes.

\subsection{Second-Order Anchor}
\label{second-order}

In this section, we first explain why the proposed second-order anchor enables the model to achieve superior image rendering with compact anchor size. The key lies in leveraging covariance-based second-order feature statistics to achieve anchor feature augmentation. Specifically, second-order feature statistics extract co-varying relationships across anchor feature dimensions. As textures and structures are defined not only by individual features but also by their interdependencies, these co-varying relationships can guide the capture of intricate local textural and structural patterns within each anchor, facilitating feature augmentation. As a result, SOGS compensates for the reduced feature dimensions and ensures superior rendering quality by performing feature augmentation. We provide the visualization of the impact of second-order anchors on scene textures and structures in the Sec.~\ref{vis}.

Given anchor feature set $\boldsymbol{F^a} \in \mathbb{R}^{N\times D} = [\boldsymbol{f_1^a}, \boldsymbol{f_2^a}, ..., \boldsymbol{f_N^a}]$, where $N$ and $D$ denote the number of anchors and the dimensionality of each anchor feature, we first compute second-order statistics in the anchor feature set. Specifically, we treat $D$ feature dimensions (channels) as $D$ variables and utilize the anchor set as $N$ observation samples, and compute the covariance matrix $\boldsymbol{\Sigma} \in \mathbb{R}^{D \times D}$  across these channels, which can be formulated as:
\begin{equation}
\boldsymbol{\Sigma} = \frac{1}{N-1} (\boldsymbol{F^a}- \boldsymbol{\mu}^T)(\boldsymbol{F^a}- \boldsymbol{\mu}^T)^T,
\end{equation}
\begin{equation}
\boldsymbol{\mu} = \frac{1}{N} \boldsymbol{F^a},
\end{equation}
where $\boldsymbol{\mu} \in \mathbb{R}^D$ denotes the mean vector across $N$ samples, used for feature centering. Given that two variables with large variances can yield a large covariance despite only weak relationship, we then construct a correlation matrix $\boldsymbol{R}$ from the covariance matrix $\boldsymbol{\Sigma}$ to standardize the relationships among $D$ variables, eliminating the influence of different scales or variances:
\begin{equation}
\boldsymbol{R} = \boldsymbol{A^{-1}\Sigma A^{-1}},
\end{equation}
\begin{equation}
\boldsymbol{A} = diag(\sigma_1, ..., \sigma_u, ..., \sigma_D), \quad \sigma_u = \sqrt{\Sigma_{uu}}
\end{equation}
where $\boldsymbol{A}$ is a diagonal matrix and the diagonal elements are the standard deviation of $D$ variables. The correlation between $u$-th and $v$-th variables can then be expressed as:
\begin{equation}
R_{uv} = \frac{\Sigma_{uv}}{\sigma_u \sigma_v},
\end{equation}
where $\Sigma_{uv}$ represents the covariance between the $u$-th and $v$-th variables. With $\boldsymbol{R}$, we extract $M$ principle co-varying relationships across anchor feature dimensions. Specifically, we perform eigendecomposition and decompose the matrix $\boldsymbol{R}$ into the eigenvector matrix $\boldsymbol{Q} \in \mathbb{R}^{D\times D}$ and the diagonal matrix $\Lambda \in \mathbb{R}^{D \times D}$ of eigenvalues:
\begin{equation}
\boldsymbol{R} = \boldsymbol{Q\Lambda Q}^T,
\end{equation}
where $\boldsymbol{Q}$ is an orthogonal matrix and each column is an eigenvector representing a co-variation pattern across anchor feature dimensions. We sort the eigenvalues in descending order, where the eigenvector corresponding to the largest eigenvalue captures the most significant co-variation patterns across the feature dimensions. Following this, we select the top-$M$ eigenvectors $\boldsymbol{P} = [\boldsymbol{P}_1, ... , \boldsymbol{P}_M]$ from $\boldsymbol{Q}$ as the primary directions of co-variation.  

We then leverage $\boldsymbol{P}$ to guide the extraction of textural and structural information in each anchor. As the co-varying relationships across anchor feature dimensions are extracted from the anchor set $\boldsymbol{F^a}$, we treat $\boldsymbol{P}$ as the global-shared co-variations. Combined with the feature $\boldsymbol{f^a}$ in one anchor, we can capture the anchor-specific textures and structures as follows:
\begin{equation}
\boldsymbol{f^t_{i}} = F_i([\boldsymbol{P}_i, \boldsymbol{f^a}]), \quad i \in [1, M],
\end{equation}
where $F_i(\cdot)$ represents the two-layer MLP used to extract the textures and structures for anchor $\boldsymbol{f^a}$ based on the $i$-th principal co-varying relationship across anchor feature dimensions. Finally, we concatenate the extracted features with the original anchor feature to achieve feature augmentation and predict Gaussian attributes. Take one second-order anchor as an example:
\begin{equation}
\{\alpha_k, c_k, q_k, s_k\}_{k=0}^{K} = F(\boldsymbol{f^a}, \{\boldsymbol{f^t_i}\}_{i=0}^M, \bm{\delta}_{ac},\boldsymbol{d}_{ac}).
\end{equation}
Note that $\{\boldsymbol{f^t_i}\}_{i=0}^M$ are derived from $\boldsymbol{f^a}$ for feature augmentation and do not increase the anchor feature dimensions and the size of each anchor.

\subsection{Selective Gradient Loss}
\label{selective_loss}
\begin{figure}[t]
\begin{center}
\includegraphics[width=1\linewidth]{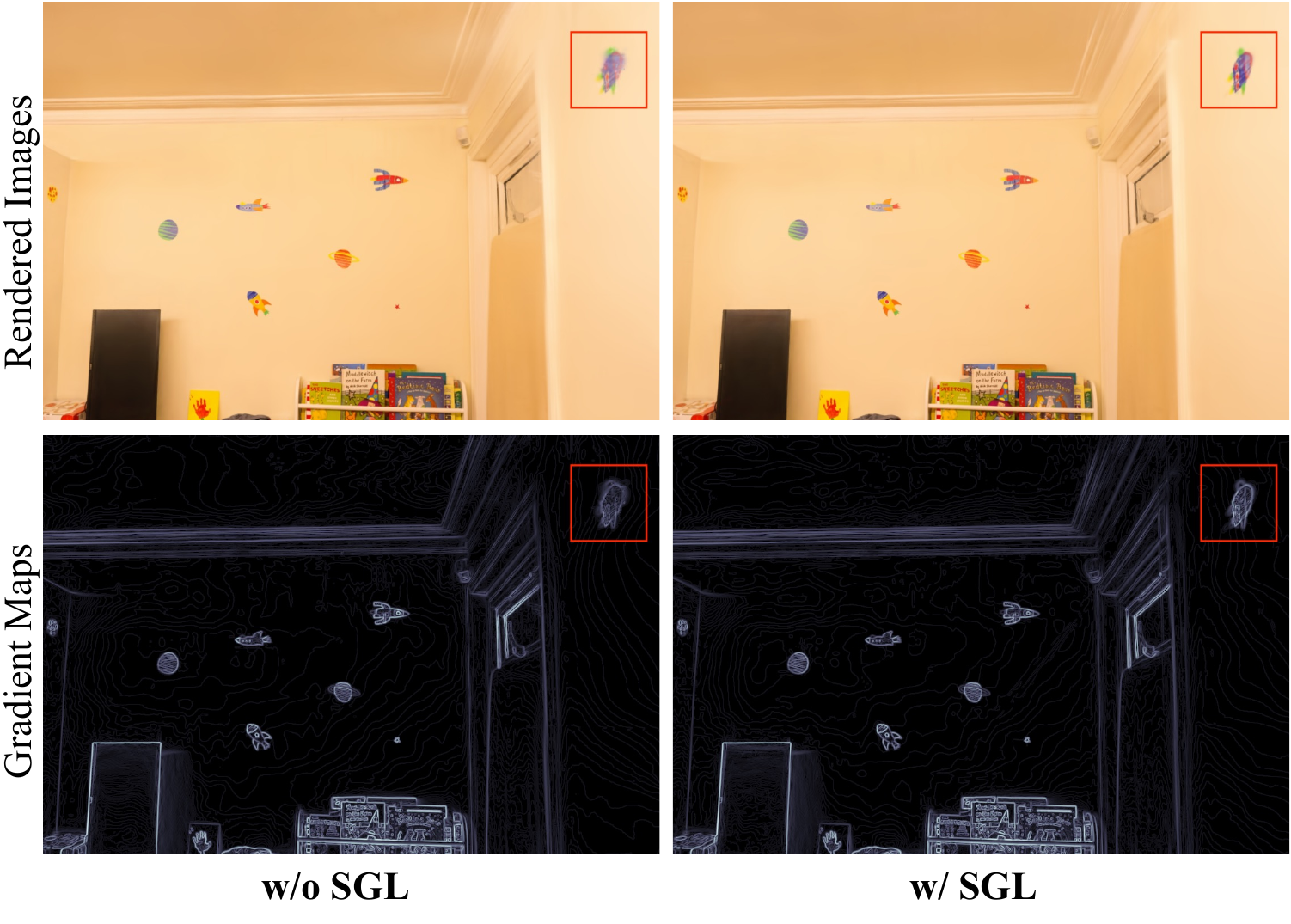}
\end{center}
\caption{
\textbf{Visual illustration of the proposed selective gradient loss (SGL).} SGL clearly improves the rendering quality by generating finer textures and details. \textbf{Zoom in for best view.}
}
\label{sgl_result}
\end{figure}

In addition to proposing second-order anchors for anchor feature augmentation to achieve superior rendering quality with reduced anchor size, we design a selective gradient loss to further ensure the rendering quality under compact anchor features. It enables the model to adaptively identify difficult-to-render textures and structures, and regularize Gaussian prediction corresponding to these regions. With the designed loss, superior rendering quality with finer textures can be achieved as shown in Fig~\ref{sgl_result}. Given the limited sensitivity of pixel-level L1 loss to textures and structures in RGB images, we consider gradient map to emphasize textural and structural regions. Specifically, we exploit the Sobel operator to extract horizontal and vertical gradient maps ($G'_x$, $G'_y$, $G_x$, $G_y$) of the rendering result $I' \in \mathbb{R}^{H \times W \times C}$ and its corresponding ground truth $I$, which can be expressed as follows:
\begin{equation}
\begin{aligned}
G'_x = S_x * I', \quad G'_y = S_y * I', \\
G_x = S_x * I, \quad G_y = S_y * I,
\end{aligned}
\end{equation}
\begin{equation}
S_x = \begin{bmatrix}
-1 & 0 & 1 \\
-2 & 0 & 2 \\
-1 & 0 & 1
\end{bmatrix}, \quad
S_y = \begin{bmatrix}
-1 & -2 & -1 \\
0 & 0 & 0 \\
1 & 2 & 1
\end{bmatrix},
\end{equation}
where $S_x$ and $S_y$ denote the horizontal and vertical Sobel kernels, respectively. We then use Euclidean metric to measure the horizontal and vertical gradient discrepancies ($l_x$, $l_y$) between the rendering $I'$ and the corresponding ground truth $I$, which are then averaged to derive the final discrepancies as follows: 
\begin{equation}
\begin{aligned}
l_x = \frac{1}{\sqrt{HW}} \sum_{x=0}^{H-1} \sum_{y=0}^{W-1} \big| G'_x(x, y) - G_x(x, y) \big|, \\
l_y = \frac{1}{\sqrt{HW}} \sum_{x=0}^{H-1} \sum_{y=0}^{W-1} \big| G'_y(x, y) - G_y(x, y) \big|.
\end{aligned}
\end{equation}

Gradient maps, derived from differences in pixel intensities, typically exhibit meaningful gradients in localized regions, such as edges and areas with significant intensity changes, which are crucial for capturing structures and textures. In contrast, the majority of areas in gradient map comprise flat and low-gradient regions with limited information. Therefore naively adopting the horizontal and vertical gradient discrepancies as the loss function can cause low-gradient regions to dominate the overall loss. As a result, the model may deemphasize regions critical for preserving fine textures and structures, focusing instead on less important flat areas. Besides, the regions with meaningful gradients should also be distinguished, prioritizing greater attention to textures and structures exhibiting larger rendering errors. To address the above issues, we introduce a dynamic region selection that computes the rendering error map as a weight map, defined by the absolute difference between the gradient maps of the rendered result and the ground truth, which can be expressed by:
\begin{equation}
\begin{aligned}
w_x = \big| G'_x(x, y) - G_x(x, y) \big|, \\
w_y = \big| G'_y(x, y) - G_y(x, y) \big|.
\end{aligned}
\end{equation}

Combined with the horizontal and vertical weight maps, the selective gradient loss $\mathcal{L}_s$ can be formulated as:
\begin{equation}
\mathcal{L}_s = w_x * l_x + w_y * l_y,
\end{equation}

Based on the above analysis, the selective gradient loss enables the model to focus on the difficult-to-render textural and structural regions adaptively. Besides, the model is able to adjust the focused regions dynamically in response to the evolving rendering error map throughout the training process. Note, the pixel-level L1 loss with D-SSIM term between RGB images is also used in the training process, which complements the proposed selective gradient loss used between gradient maps. And we also keep the volume regularization as described in Scaffold-GS~\cite{lu2024scaffold}. The total loss function is given as follows:
\begin{equation}
\mathcal{L} = \lambda_1\mathcal{L}_1 + \lambda_{SSIM}\mathcal{L}_{SSIM} + \lambda_{vol}\mathcal{L}_{vol} + \lambda_s\mathcal{L}_s,
\end{equation}
where the $\lambda_s$, $\lambda_1$, $\lambda_{SSIM}$ and $\lambda_{vol}$ represent the training weights of the selective gradient loss $\mathcal{L}_s$, the L1 loss $\mathcal{L}_1$, the D-SSIM term $\mathcal{L}_{SSIM}$ and the volume regularization $\mathcal{L}_{vol}$.
\begin{table*}[t]
\definecolor{red}{rgb}{1,0.6,0.6}
\definecolor{orange}{rgb}{1,0.8,0.6}
\definecolor{yellow}{rgb}{1,1,0.6}
        \renewcommand\arraystretch{1.5}
	\renewcommand\tabcolsep{1.1pt}
		\begin{tabular}{l|cccc|cccc|cccc}
                \hline
                
			Datasets & \multicolumn{4}{c|}{Mip-NeRF360}  & \multicolumn{4}{c|}{Tanks\&Temples} & \multicolumn{4}{c}{Deep Blending}\\
                \hline
			Methods
			& SSIM$^\uparrow$   & PSNR$^\uparrow$    & LPIPS$^\downarrow$  & Anchor Size
			& SSIM$^\uparrow$   & PSNR$^\uparrow$    & LPIPS$^\downarrow$  & Anchor Size
			& SSIM$^\uparrow$   & PSNR$^\uparrow$    & LPIPS$^\downarrow$  & Anchor Size \\
			\hline 
                Scaffold-GS & 0.806 & 27.50 & 0.252 & 32 dim & 0.853 & 23.96 & 0.177 & 32 dim & 0.906 & 30.21 & 0.254 & 32 dim \\
                \hline
                SOGS(Ours) & \textbf{0.815} & \textbf{27.85} &\textbf{ 0.221} & \textbf{16 dim} & \textbf{0.855} & \textbf{24.14} & \textbf{0.176} & \textbf{12 dim} & \textbf{0.907} & \textbf{30.29} & \textbf{0.252} & \textbf{12 dim} \\
                \hline
			
		\end{tabular}
	\caption{\textbf{Quantitative comparisons} of novel view synthesis. All methods are trained with the same training data. SOGS achieves superior rendering quality with reduced anchor (model) size. Notably, SOGS using the feature dimension of 12 can still outperform the Scaffold-GS that uses a feature dimension of 32. We do not compare with HAC~\cite{chen2024hac} and ContextGS~\cite{wang2024contextgs}, as they leverage compression techniques to reduce the storage size of Scaffold-GS rather than the actual anchor and model sizes as used in training and rendering.
	}
 \label{quantitative_result}
\end{table*}

\begin{table}[t]
\definecolor{red}{rgb}{1,0.6,0.6}
\definecolor{orange}{rgb}{1,0.8,0.6}
\definecolor{yellow}{rgb}{1,1,0.6}
        \renewcommand\arraystretch{1.3}
	\renewcommand\tabcolsep{4.4pt}
		\begin{tabular}{l|cccc}
                \hline
                
			Datasets & \multicolumn{4}{c}{BungeeNeRF} \\
                \hline
			Methods
			& SSIM$^\uparrow$   & PSNR$^\uparrow$    & LPIPS$^\downarrow$  & Anchor Size \\
			\hline 
                Scaffold-GS & 0.865 & 26.62 & 0.241 & 32 dim \\
                \hline
                SOGS(Ours) & \textbf{0.880} & \textbf{27.06} & \textbf{0.171} & \textbf{16 dim} \\
                \hline
			
		\end{tabular}
	\caption{\textbf{Quantitative comparisons} on large-scale scenes from the dataset BungeeNeRF~\cite{xiangli2022bungeenerf}. All methods are trained with the same training data. 
	}
 \label{large_scale_result}
\end{table}

\section{Experiments}

\subsection{Datasets and Implementation Details}
\label{dataset}
\paragraph{Datasets} For both training and testing, we adopt the datasets and configurations in Scaffold-GS~\cite{lu2024scaffold} and conduct experiments on images of 19 real-world scenes. Specifically, we evaluate SOGS across all nine scenes from the Mip-NeRF360 dataset \cite{barron2022mip}, two scenes from the Tanks\&Temples dataset \cite{knapitsch2017tanks}, and two additional scenes from the Deep Blending dataset \cite{hedman2018deep}. These scenes exhibit diverse styles, ranging from bounded indoor environments to unbounded outdoor ones. We also evaluate SOGS on the BungeeNeRF dataset~\cite{xiangli2022bungeenerf}, which includes six large-scale outdoor scenes, to further validate its effectiveness. The datasets are divided into training and test sets following the same settings as Scaffold-GS, with image resolutions consistent with those used in Scaffold-GS.

\begin{figure*}[t]
\begin{center}
\includegraphics[width=1\linewidth]{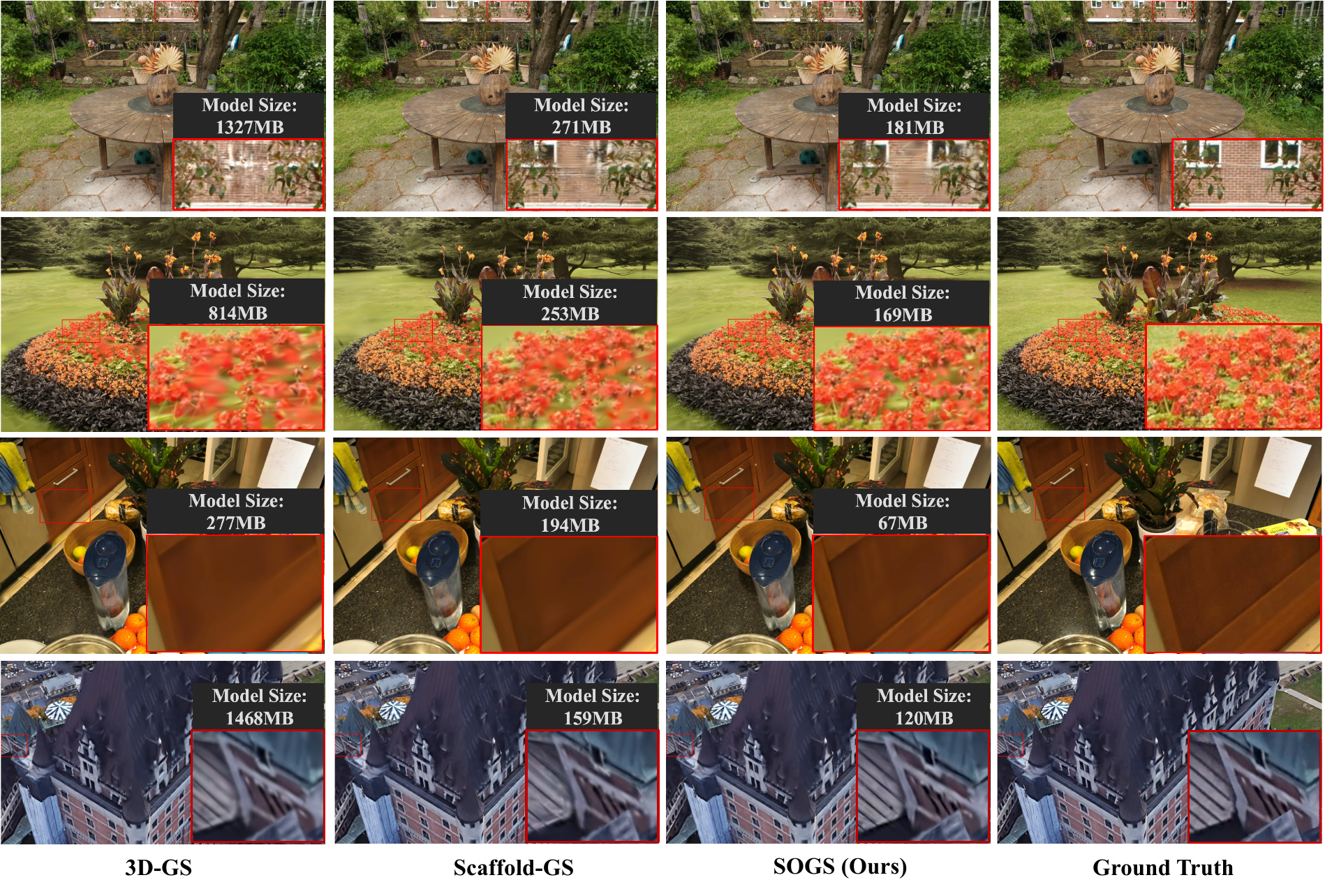}
\end{center}
\caption{
\textbf{Qualitative comparisons of SOGS with 3D-GS~\cite{kerbl20233d} and Scaffold-GS~\cite{lu2024scaffold}.} SOGS achieves smaller model size and superior image rendering with much less artifacts but more fine details. The experiments are conducted over multiple indoor and outdoor scenes including `Garden', `Flower' and `Counter' from the Mip-NeRF360 and `Quebec' from the BungeeNeRF. \textbf{Zoom in for best view.}}
\label{qualitative_comparison}
\end{figure*}

\paragraph{Implementation} For the second-order anchor, we reduce the dimension $D$ of the anchor features in Scaffold-GS from 32 to 16, or even to 12. The covariance matrix $\boldsymbol{\Sigma}$ is computed from all anchors in scene representations which captures the global correlation across all anchor feature dimensions. The number of the selected eigenvectors is set to $M=2$. All MLPs used to extract anchor-specific textural and structural information consist of two layers with ReLU activation. For the loss function, the training weights $\lambda_s$, $\lambda_1$, $\lambda_{SSIM}$ and $\lambda_{vol}$ are set to $0.01$, $0.8$, $0.2$ and $0.01$. 
Please note, for fair comparisons, we keep the loss terms $\lambda_1$, $\lambda_{SSIM}$ and $\lambda_{vol}$ the same as in Scaffold-GS. 
The number of 3D Gaussians corresponding to one second-order anchor is set as $K=10$. We use the Pytorch framework to implement SOGS. The SOGS model is trained for 30000 iterations, the same as Scaffold-GS and 3D-GS.

\subsection{Comparisons with the State-of-the-Art}

We primarily compare SOGS with Scaffold-GS~\cite{lu2024scaffold} over datasets Mip-NeRF360, Tank\&Temple and Deep Blending as Scaffold-GS is an anchor-based 3D Gaussian splatting method and is most relevant to our study. Table~\ref{quantitative_result} shows experimental results over the same test images as described in Section~\ref{dataset}. We can observe that SOGS outperforms Scaffold-GS consistently in PSNR, SSIM and LPIPS across all evaluated scenes. The anchor size of SOGS is reduced to 16 in the dataset Mip-NeRF360, while it is challengingly reduced to 12 in Tank\&Temple and Deep Blending. The superior performance is largely attributed to our proposed second-order anchor which performs anchor feature augmentation to compensate for the reduced anchor feature size and improve the rendering quality as well as the selective gradient loss that further ensures the high-quality rendering under small-sized anchor features. Further, we benchmark SOGS with Scaffold-GS over the BungeeNeRF dataset which provides diverse large-scale scenes. It can be observed that SOGS achieves superior rendering quality with only 16-dimensional anchors. Besides, as shown in Fig.~\ref{qualitative_comparison}, we perform the visual comparison of SOGS with 3D-GS and Scaffold-GS and provide their model sizes in each scene. With the compact anchor size of the proposed second-order anchor, SOGS achieves a significant reduction in model size while delivering superior novel view synthesis with much less artifacts but finer details. 
Note we did not compare with HAC~\cite{chen2024hac} and ContextGS~\cite{wang2024contextgs} as they compress the storage size of Scaffold-GS rather than the actual model size as used in training and rendering.
\subsection{Ablation Studies}

\renewcommand\arraystretch{1.1}
\begin{table}[t]
\renewcommand\tabcolsep{7.5pt}
\begin{center}
    
\begin{tabular}{l||ccc} 
\hline
& 
\multicolumn{3}{c}{Evaluation Metrics}
\\
\cline{2-4}
\multirow{-2}{*}{Models} 
& PSNR $\uparrow$ 
& SSIM $\uparrow$
& LPIPS $\downarrow$ 
\\\hline
Base & 26.62 & 0.865 & 0.241 \\

Base+SOA & 27.25 & 0.879 & 0.208 \\

Base+SOA+SGL & \textbf{27.39} & \textbf{0.887} & \textbf{0.161} \\

\hline
\end{tabular}
\end{center} 
\caption{
\textbf{
Ablation studies of SOGS} on the dataset BungeeNeRF. With Scaffold-GS as the baseline \textit{Base}, \textit{Base+SOA} introduces the second-order anchor to augment anchor features which clearly improves the Gaussian attribute prediction and image rendering. \textit{Base+SOA+SGL} (i.e., SOGS) further introduces the selective gradient loss which enables the Gaussian prediction to focus on regions with difficult-to-render textures and geometries. All models are trained under the same settings and model size, and the anchor feature dimension is set at 32.
}
\label{ablation}

\end{table}

We conduct extensive ablation experiments to validate the effectiveness of the proposed second-order anchor and the selective gradient loss. In the experiments, we keep the model size and anchor feature dimension the same as used in Scaffold-GS, aiming to screen out other variations and show how our two designs improve the rendering quality. More details are described in the following two subsections.

\paragraph{Second-order Anchor} We first examine how our proposed second-order anchor affects PSNR, SSIM and LPIPS. In this experiment, we adopt the Scaffold-GS as the baseline model \textit{Base} that performs anchor-based Gaussian splatting with naive anchors. On top of the \textit{Base}, we train a model \textit{Base+SOA} that incorporates our proposed second-order anchor to replace the naive anchor in Scaffold-GS. As Table \ref{ablation} shows, \textit{Base+SOA} outperforms the \textit{Base} clearly in PSNR, SSIM and LPIPS, indicating that the second-order anchor can effectively improve the Gaussian attribute prediction and novel view synthesis by anchor feature augmentation.

\paragraph{Selective Gradient Loss} To evaluate how the proposed selective gradient loss contributes, we train a new model \textit{Base+SOA+SGL} (i.e., the complete SOGS) that incorporates the selective gradient loss on top of \textit{Base+SOA}. Table \ref{ablation} shows experimental results. We can observe that \textit{Base+SOA+SGL} further improves the novel view synthesis consistently across all three evaluation metrics, demonstrating its effectiveness on Gaussian attribute prediction and rendering quality.

\begin{figure}[t]
\begin{center}
\includegraphics[width=1\columnwidth]{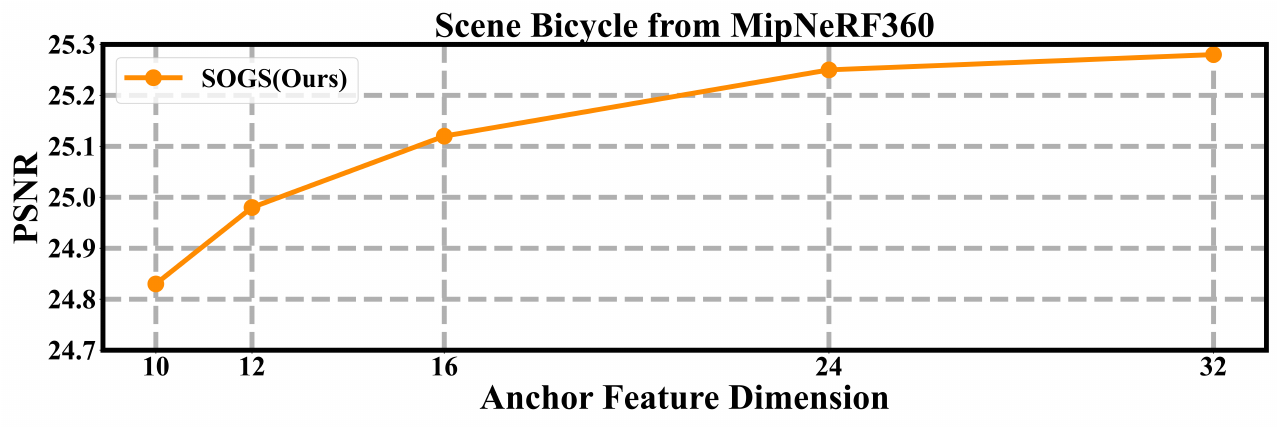}
\end{center}
\caption{
\textbf{Anchor feature dimension vs SOGS performance:} The performance of SOGS varies with the feature dimension $D$ in each anchor, where increasing $D$ improves SOGS consistently together with the increased model size and computational cost. The graph shows the PSNR of the scene `Bicycle' from  MipNeRF360.
}
\label{para}
\end{figure}

\subsection{Parameter Investigation}

We examine how the dimension of anchor features $D$ affects the SOGS performance. Fig.~\ref{para} shows experiments on the scene `Bicycle' from the dataset MipNeRF360. We can observe that the performance of SOGS (in PSNR $\uparrow$ score) improves consistently while the feature dimension $D$ increases. Specifically, the feature dimension plays a vital role in the rendering quality as it sets the upper bound for the encoded information as well as the effectiveness of feature augmentation in SOGS. As $D$ becomes larger than 16, the performance gains gradually diminish due to feature saturation while the model size and computational costs grow greatly. We therefore set $D$ at 12 or 16 in our implemented system to balance the model size and performance.

\begin{figure}[t]
\begin{center}
\includegraphics[width=0.96\columnwidth]{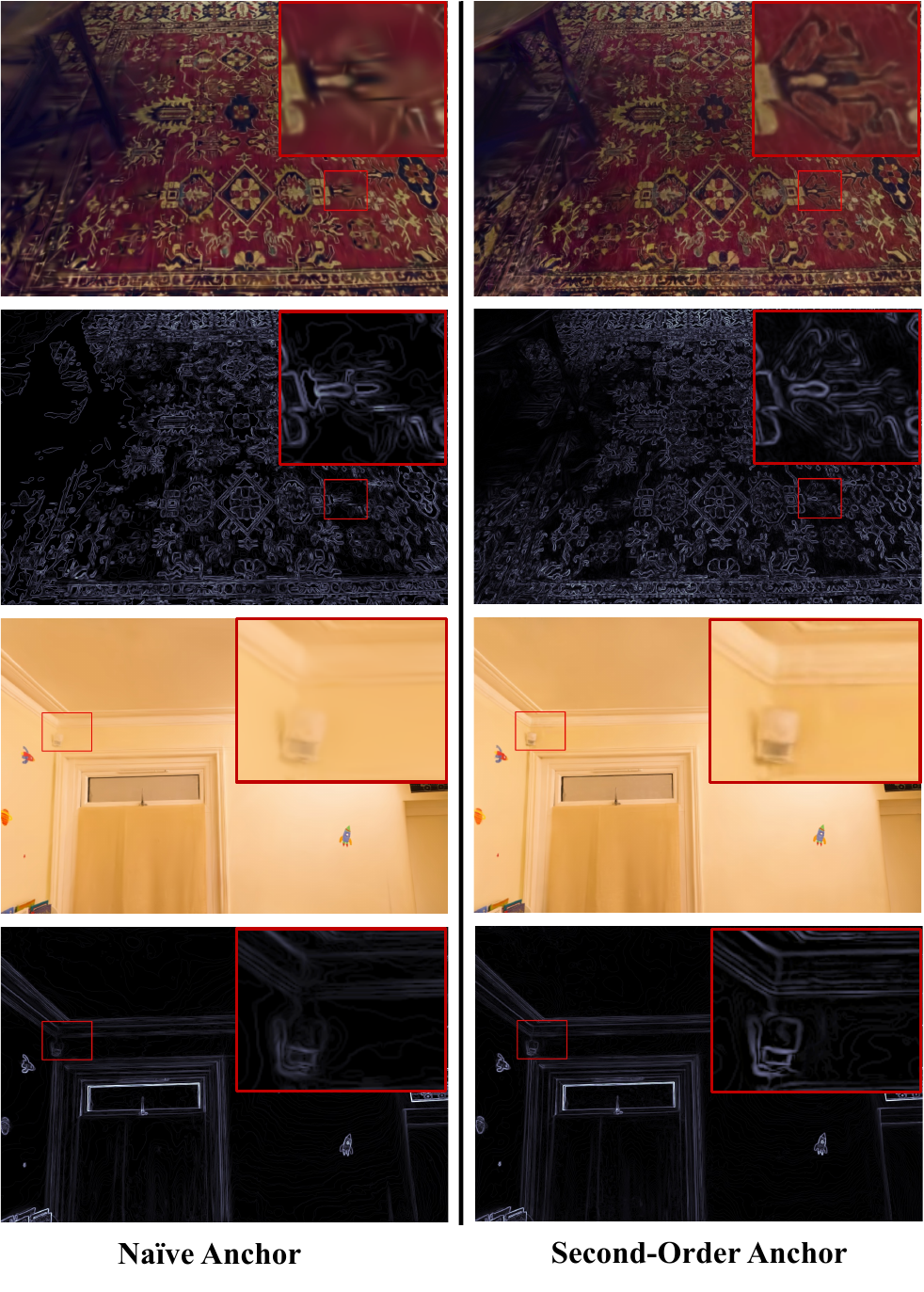}
\end{center}
\caption{\textbf{Visualization of the impact of second-order anchors on scene textures and structures.} The two samples are `Drjohnson' and `Playroom' from dataset Deep Blending~\cite{hedman2018deep}. The proposed second-order anchor improves the rendering quality with superior structures and textures by anchor feature augmentation. \textbf{Zoom in for best view.}}
\label{sogs_visualization}
\end{figure}

\subsection{Visualization}
\label{vis}

We visualize the impact of second-order anchors on scene textures and structures. As Fig.~\ref{sogs_visualization} shows, our proposed second-order anchor can enhance the rendered images with finer structures and textures. This visualization verifies that the proposed second-order anchor can effectively capture the structural and textural patterns for anchor feature augmentation, and facilitate subsequent Gaussian prediction and image rendering with augmented features.
\section{Conclusion}
This paper presents SOGS, an innovative anchor-based 3D Gaussian splatting technique that incorporates second-order anchors to achieve superior rendering with compact anchor feature dimensions and model size. Specifically, by introducing covariance-based second-order statistics, SOGS learns correlations across anchor feature dimensions to capture anchor-specific textures and structural patterns. This enables each anchor to achieve feature augmentation, compensating for the reduced anchor feature size and resulting in superior Gaussian prediction and rendering quality. Besides, we also design a selective gradient loss to further ensure high-quality rendering under compact anchor features, which enables the model to dynamically focus on the Gaussian prediction for difficult-to-render textural and structural regions.
Experiments over multiple widely adopted indoor and outdoor scenes show that SOGS achieves superior novel view synthesis with compact model size.
Although time complexity increases slightly due to the additional computations in second-order anchors, SOGS still achieves efficient training and real-time rendering.

\section{Acknowledgements}

This project is funded by the Ministry of Education Singapore, under the Tier-2 project scheme with a project number MOE-T2EP20220-0003.
{
    \newpage
    \small
    \bibliographystyle{ieeenat_fullname}
    \bibliography{main}
}

\end{document}